\documentclass[sw]{iosart2c}
\pdfoutput=1

\usepackage{natbib}
\usepackage{times}
\usepackage{amssymb}
\usepackage{amsfonts}
\usepackage{dcolumn}
\usepackage{graphicx}
\usepackage{vntex}
\usepackage[english]{babel}
\usepackage{balance}
\usepackage[bottom]{footmisc}

\newcolumntype{d}[1]{D{.}{.}{#1}}

\usepackage[skip=1pt]{caption}

\usepackage{url}
\usepackage[hidelinks]{hyperref}
\usepackage{fixltx2e}

\begin{document}
\begin{frontmatter}

\title{Ripple Down Rules for Question Answering}

\runningtitle{Ripple Down Rules for Question Answering}

\review{Christina Unger, Bielefeld University, Germany; Axel-Cyrille Ngonga Ngomo, University of Leipzig, Germany; Philipp Cimiano, Bielefeld University, Germany; S{\"o}ren Auer, University of Bonn, Germany; George Paliouras, NCSR Demokritos, Greece}{Gosse Bouma, University of Groningen, Netherlands; Konrad H{\"o}ffner, University of Leipzig, Germany; Shizhu He, Chinese Academy of Sciences, China; Christina Unger, Bielefeld University, Germany}{}

\author[A]{\fnms{Dat Quoc} \snm{Nguyen}\thanks{The first two authors contributed equally to this work. Corresponding author's e-mail: dat.nguyen@students.mq.edu.au.}},
\author[B]{\fnms{Dai Quoc} \snm{Nguyen}}
and
\author[C]{\fnms{Son Bao} \snm{Pham}}
\runningauthor{Nguyen et al.}
\address[A]{Department of Computing,    Macquarie University, Australia \\
E-mail: dat.nguyen@students.mq.edu.au}
\address[B]{Department of Computational Linguistics, Saarland University, Germany \\
E-mail: daiquocn@coli.uni-saarland.de}
\address[C]{VNU University of Engineering and Technology, Vietnam National University, Hanoi, Vietnam\\
E-mail: sonpb@vnu.edu.vn}

\begin{abstract}
\  Recent years have witnessed a new trend of building ontology-based question answering systems. These systems use semantic web information to produce more precise answers to users' queries. 
However, these systems are mostly designed for English. In this paper, we introduce an ontology-based question answering system named KbQAS which, to the best of our knowledge, is the first one made for Vietnamese.
KbQAS employs our question analysis approach that systematically constructs a knowledge base of grammar rules to convert  each input question into an intermediate representation element.
KbQAS then takes the intermediate representation element with respect to a target ontology and applies concept-matching techniques to return an answer.
On a wide range of Vietnamese questions, experimental results show that the performance of  KbQAS   is promising with accuracies of 84.1\% and 82.4\%  for analyzing input questions and retrieving output answers, respectively. 
Furthermore, our question analysis approach can easily be applied to new domains and new languages, thus saving time and human effort.
\end{abstract}


%\vspace{-20pt}

\end{frontmatter}

\section{Introduction}
\label{sec:Intro}

Accessing online resources often requires the support from  advanced information retrieval technologies  to produce expected information. 
 This brings new challenges to the construction of  information retrieval systems such as search engines and question answering (QA) systems. Given an input query expressed in a  keyword-based mechanism, most search engines return a long list of title and short snippet pairs  ranked by their relevance to the input query. Then the user has to scan the list to get the expected information, so this is a time consuming task \cite{Zeng2004}. Unlike  search engines,  QA systems directly produce an exact answer to an input question.  In addition,  QA systems allow to specify the input question in natural language rather than as keywords. 
 
In general, an open-domain QA system aims to potentially answer any  user's question. In contrast, a restricted-domain QA system only handles the questions related to a specific domain. Specifically,  traditional restricted-domain QA systems make use of  relational databases to represent target domains. Subsequently, with the advantages of the semantic web, the recent restricted-domain QA systems employ knowledge bases such as ontologies as the target domains \cite{McGuinness2004}. Thus, semantic markups can be used to add meta-information to return precise answers for complex natural language questions. This is an avenue which has not been actively explored for Vietnamese.

In this paper, we introduce the first ontology-based QA system for Vietnamese, which we call KbQAS. KbQAS consists of question analysis and answer retrieval components. The  question analysis component uses a knowledge base of grammar rules for analyzing input questions; and the   answer retrieval component is responsible for interpreting the input questions with respect to a target ontology.
The association between the two components is an intermediate representation element which captures the semantic structure of any input question. This intermediate element contains properties of the input question including question structure, question category, keywords and semantic constraints between the keywords.

The {\textit{key innovation}} of KbQAS is that it proposes a knowledge acquisition approach to systematically build a knowledge base for analyzing natural language questions.
To convert a natural language question into an explicit representation in a QA system, most previous works so far have used rule-based approaches, to the best of our knowledge. The manual creation of rules in an ad-hoc manner is more expensive in terms of time and effort, and it is error-prone because of the representation complexity and the variety of structure types of the questions.
For example,  rule-based methods, such as for English \cite{LopezUMP07} and for Vietnamese as described in the first KbQAS version \cite{NguyenNP09},  manually define a list of  pattern structures to analyze the questions.
As rules are created in an ad-hoc manner, these methods share common difficulties in controlling the interaction between the rules and keeping the consistency among them.
In our question analysis approach, however, we apply Single Classification Ripple Down Rules knowledge acquisition methodology \cite{ComptonJ90,RichardsD09} to acquire the rules in a systematic manner, where  consistency between rules is maintained and an unintended interaction among rules is avoided.
Our approach allows an easy adaptation to a new domain and a new language and saves time and effort of human experts.

The paper is organized as follows. We provide  related work in Section \ref{sec:relatedworks}. We describe  KbQAS and our knowledge acquisition approach for question analysis in Section \ref{sec:VnQAS} and Section \ref{sec:rdrqa}, respectively. We evaluate KbQAS in Section \ref{sec:experiments}. The conclusion will be presented in Section \ref{sec:conclusion}.

\section{Short overview of question answering}
\label{sec:relatedworks}

\subsection{Open-domain question answering}

The goal of an open-domain QA system is to automatically return an answer for every natural language question \cite{HirschmanG01,Webber2010,Mendes2012}. For example, such systems as START \cite{Katz97}, FAQFinder \cite{BurkeHKLTS97} and AnswerBus \cite{Zheng2002} answer questions over the Web.  
 Subsequently, the question paraphrase recognition task is considered as one of the important tasks in QA. Many proposed  approaches for this task are  based on machine learning as well as knowledge representation and reasoning  \cite{Bernhard2008,Jijkoun05,Rinaldi2003,Zhao07ijcai,fader-zettlemoyer-etzioni:2013:ACL2013,berant-liang:2014:P14-1}.

Since aroused by the QA track of the Text Retrieval Conference \cite{Voorhees2001} and the multilingual QA track of the CLEF conference \cite{Penas2012}, many open-domain QA systems from the information retrieval perspective \cite{Kolomiyets2011} have been introduced. 
For example, in the TREC-9 QA competition \cite{Voorhees00}, the Falcon system \cite{HarabagiuMPMSBGRM00}  achieved the highest results. The innovation of  Falcon  focused on a method using   WordNet \cite{FellbaumC98} to boost its knowledge base. In the QA track of the TREC-2002 conference \cite{Voorhees02overviewof}, the PowerAnswer system \cite{MoldovanHGMLABB02} was the most powerful system, using a deep linguistic analysis. 

\subsection{Traditional restricted-domain question answering}

Usually linked to relational databases,  traditional restricted-domain QA systems are called natural language interfaces to databases. A natural language interface to a database (NLIDB) is a system that allows the users to access information stored in a database by typing questions using natural language expressions \cite{AndroutsopoulosRT95}.  
 In general, NLIDB systems focus on converting the input question into an expression in the corresponding database query language.
 For example, the LUNAR system \cite{WoodsKW72} transfers the input question into a parsed tree, and the tree is then  directly converted into an expression in a database query language.   However, it is difficult to create converting rules that directly transform the tree into the query expression. 

Later NLIDBs, such  as  Planes \cite{Waltz78},  Eufid \cite{TempletonB83}, PRECISE \cite{PopescuEK03}, C-Phrase \cite{Minock10} and the systems presented in \cite{StraticaKD03,NguyenL08},  use semantic grammars to analyze questions. The semantic grammars consist of the hard-wired knowledge orienting a specific domain, so these NLIDB systems need to develop new grammars whenever porting to a new knowledge domain.

Furthermore, some systems, such as  TEAM \cite{Martin1986} and MASQUE/SQL \cite{AndroutsopoulosRT93}, use  syntactic-semantic interpretation rules driving logical forms to process the input question. 
 These systems firstly transform the input question into an intermediate logical expression of high-level world concepts without any relation to the database structure. The logical expression is then converted to an expression in the database query language.   Here, using the logical forms enables those systems to  adapt to other domains as well as to different query languages \cite{SilakariMN11}. In addition, there are many systems also using logical forms to process the input question, e.g.  \cite{thompson:ml97ws,MoldovanHGMLABB02,BenjaminHKN03,Furbach2010,Dong2011,Liu2012,berant-EtAl:2013:EMNLP}.

\subsection{Ontology-based question answering}

As a knowledge representation of a set of concepts and their relations in a specific domain, an ontology can provide semantic information to handle  ambiguities, to interpret and answer user questions in terms of QA \cite{Lopez2011}. A discussion on the construction approach of an ontology-based QA system can be found in \cite{basiliEtAl2004}. This approach was then applied to build the MOSES system \cite{AtzeniBHMPPZ04}, with the focus on the question analysis. The following systems are some typical ontology-based QA systems.

The AquaLog system \cite{LopezUMP07} performs  semantic and syntactic analysis of the input question using resources including word segmentation, sentence segmentation and part-of-speech tagging,  provided by the GATE framework \cite{CunninghamMBT02}. When a question is asked, AquaLog transfers the  question into a query-triple form of ({generic term, relation, second term}) containing the keyword concepts and relations in the question, using JAPE grammars in GATE. AquaLog then matches  each element in the query-triple to an element in the target ontology to create an onto-triple, using string-based comparison methods and WordNet \cite{FellbaumC98}. 
Evolved from  AquaLog, the PowerAqua system \cite{LopezFMS12} is an  open-domain system, combining the knowledge from various heterogeneous ontologies which were autonomously created on the semantic web. Meanwhile,  the PANTO  system \cite{WangXZY2007} relies on the statistical Stanford parser to map an input question into a query-triple; the query-triple is then translated into an onto-triple with the help of a lexicon of all entities  from a given target ontology enlarged with WordNet synonyms; finally, the onto-triple and potential words derived from the parse tree are used to produce a SPARQL query on the target ontology. 

Using the gazetteers in the GATE framework, the QuestIO system \cite{DamljanovicTB08}  identifies the keyword concepts in an input question. Then QuestIO retrieves potential relations between the concepts  before ranking these relations based on their similarity, distance and specificity scores; and so  QuestIO creates formal SeRQL or SPARQL queries based on the concepts and the ranked relations. Later the FREyA system \cite{Damljanovic2010}, the successor of QuestIO,  allows users to enter questions in any form and interacts with the users to handle  ambiguities  if necessary.  

In the ORAKEL system \cite{CimianoHHMS08}, wh-questions are converted to F-Logic or SPARQL queries by using domain-specific Logical Description Grammars. Although ORAKEL supports compositional semantic constructions and obtains a promising performance, it involves a customization process of the domain-specific lexicon. Also, another interesting work over linked data  as detailed in \cite{Unger2012} proposed an approach to convert the syntactic-semantic representations of the input questions into the SPARQL templates. Furthermore, the Pythia system \cite{Unger2011} relies on  ontology-based grammars  to process  complex questions. However, Pythia requires a manually created lexicon. 

\subsection{Question answering and question analysis for Vietnamese}

Turning to Vietnamese question answering, \citet{NguyenL08} introduced a
Vietnamese NLIDB  system using
semantic grammars. Their system includes two main modules: 
the query translator (QTRAN) and the text generator (TGEN). QTRAN  maps
an input natural language question to an SQL query, while TGEN
 generates an answer based on the table result
 of the SQL query. The QTRAN module uses  limited context-free grammars to convert the input question into a syntax tree by means of the CYK algorithm \cite{Younger1967}. The
syntax tree is then converted into an SQL query by using a
 dictionary to identify names of attributes in the database and names of individuals
stored in these attributes. The TGEN module combines pattern-based
and keyword-based approaches to make sense of the
meta-data and relations in database tables to produce the answer.   

In our first KbQAS conference publication \cite{NguyenNP09}, we reported a hard-wired approach  to convert input  questions into  intermediate representation elements 
which are then used to extract the corresponding elements from a target ontology to return answers. Later, \citet{PhanN10}  described a method to map Vietnamese questions  into triple-like formats  ({Subject}, {Verb}, {Object}). 
 Subsequently, \citet{NguyenN2011} presented another ontology-based  QA system for Vietnamese, where keywords in an input question are identified by using pre-defined templates, and these keywords are then used to produce a SPARQL query to retrieve a triple-based answer from a target ontology.  In addition, \citet{tran-EtAl:2012:PACLIC} described the VPQA system to answer person name-related questions while \citet{NguyenHP12} presented another  NLIDB system to answer  questions in the economic survey domain. 

%%%%%%%%%%%%%%%%%%%%%%%%%%%%%%%%%%%%%%%%%%%%%%%%%%%%%%%%%%%%%%%%%%%%%%%%%%%%%%%%%%%%%%%%%%%5555

\section{Our KbQAS question answering system }
\label{sec:VnQAS}

\begin{figure}[ht]
\centering
\includegraphics[width=7.5cm]{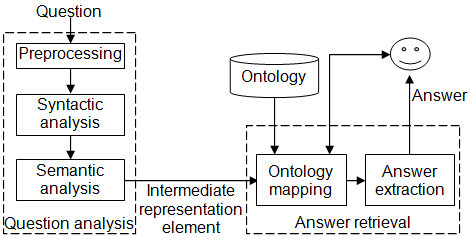}
\caption{System architecture of KbQAS.}
\label{Figure1}
\end{figure}

This section gives an overview of KbQAS. The architecture of KbQAS, as shown in Figure \ref{Figure1}, contains two components: the natural language question analysis engine and the answer retrieval component. 

%%%%%%%%%%%%%%%%%%%%%%%%%%%%%%%%%%%%%%%%%%%%%%%%%%%%%%%%%%%%%%%%%%%%%%%%%%%

The question analysis component consists of three modules: preprocessing, syntactic analysis and semantic analysis. 
This component takes the user question as an input and returns an intermediate element representing the input question in a compact form.  
The role of the intermediate representation element is to provide the structured information about the input question for the later process of answer retrieval. 

The answer retrieval component contains two modules: ontology mapping and answer extraction. It takes the intermediate representation element produced by the question analysis component and an ontology as its input to generate the answer.

%%%%%%%%%%%%%%%%%%%%%%%%%%%%%%%%%%%%%%%%%%%%%%%%%%%%%%%%%%%%%%%%%%%%%%%%%%%%%%%%%%%%%%%

\begin{figure*}[t]
\centering
\includegraphics[width=16.125cm]{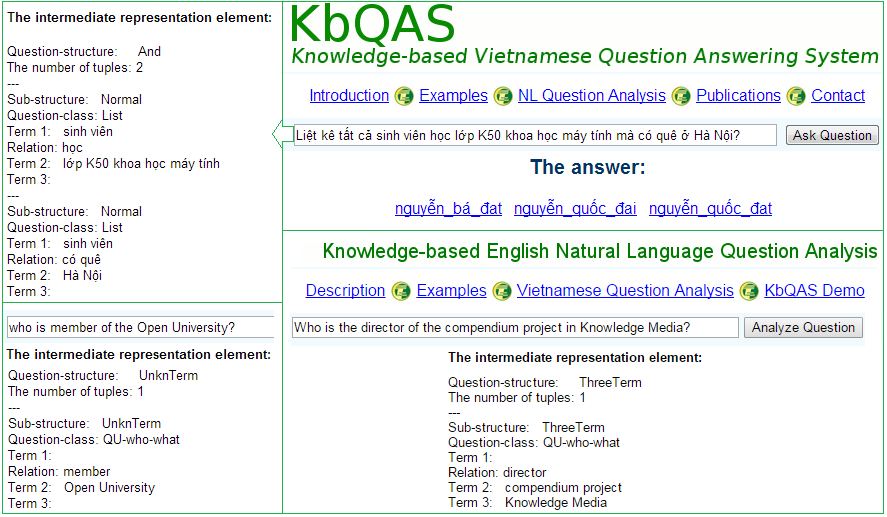}
\caption{Illustrations of question analysis and question answering.}
\label{fig:kbqa}
\end{figure*}

\subsection{Intermediate representation of an input question}

Unlike AquaLog \cite{LopezUMP07}, the intermediate representation element in KbQAS covers a wider variety of question types. 
This element consists of a {question structure} and one or more {query tuple}s in the following format:

\vspace{3pt}
{(sub-structure, question category, \emph{Term}$_1$, \emph{Relation}, \emph{Term}$_2$, \emph{Term}$_3$)}
\vspace{3pt}

\noindent where {\emph{Term}$_1$} represents a concept (i.e. an object class),  excluding the cases of  \textit{Affirm}, \textit{Affirm\_3Term} and \textit{Affirm\_MoreTuples} question structures. In addition, {\emph{Term}$_2$} and {\emph{Term}$_3$} represent entities (i.e. objects or instances), excluding the cases of \textit{Definition} and \textit{Compare} question structures. Furthermore, \emph{Relation}  is a semantic constraint between the terms. 

%This representation is aimed to capture the semantic of question.
We define the following question structures: \textit{Normal, UnknTerm, UnknRel, Definition, Compare, ThreeTerm, Clause, Combine, And, Or,  Affirm\_MoreTuples, Affirm, Affirm\_3Term}, and question categories: \textit{What, When, Where, Who, HowWhy, YesNo, Many, ManyClass, List} and \textit{Entity}. See  Appendix \textbf{A} and Appendix \textbf{B} for details of these definitions.

A simple question  has only one {query tuple} and its {question structure} is the sub-structure in the query tuple. A complex question, such as a composite one, has several sub-questions, where each sub-question is represented by a separate query tuple, and the {question structure} captures this composite factor. 

For example, the question 
{``Phạm Đức Đăng học trường đại học nào và  được hướng dẫn bởi ai ?''} (``Which university does Pham Duc Dang enroll in and who tutors him ?'') has the \textit{Or} question structure and two query tuples where ? represents a missing attribute:  ({\textit{Normal}, \textit{Entity}, trường đại học\textsubscript{university}, học\textsubscript{enroll}, Phạm Đức Đăng\textsubscript{Pham\ Duc\ Dang}, ?})
 and  
({\textit{UnknTerm}, \textit{Who}, ?, hướng dẫn\textsubscript{tutor}, Phạm Đức Đăng \textsubscript{Pham\ Duc \ Dang}, ?}).

The intermediate representation element is designed so that it can represent
various types of question structures. Therefore, attributes such as \emph{Relation} or terms
in the query tuple can be missing. For example, a question has the \textit{Normal}  question structure if it has only one query tuple and \emph{Term}$_3$ is missing.

\subsection{An illustrative example}
\label{subsec:illex}

For demonstration\footnote{The KbQAS is available at \url{http://150.65.242.39:8080/KbQAS/} with an intro video on YouTube at \url{http://youtu.be/M1PHvJvv1Z8}.}  \cite{NguyenNP13} and evaluation purposes, we reuse an ontology which models the organizational system of the VNU University of Engineering and Technology, Vietnam. 
The ontology contains 15 concepts such as ``trường\textsubscript{school}'', ``giảng viên\textsubscript{lecturer}'' and ``sinh viên\textsubscript{student}'', 17 relations or properties such as ``học\textsubscript{enroll}'', ``giảng dạy\textsubscript{teach}'' and ``là sinh viên của \textsubscript{is  student  of}'', and 78 instances, as described in our first KbQAS version \cite{NguyenNP09}.

Given a complex-structure question {``Liệt kê tất cả sinh viên học lớp K50 khoa học máy tính mà có quê ở Hà Nội''}
 (``List all students enrolled in the
K50 computer science course, whose hometown is Hanoi''), the question analysis component determines that this question has the \textit{And} question structure with two query tuples ({\textit{Normal}, \textit{List}, sinh viên\textsubscript{student}, học\textsubscript{enrolled}, lớp K50 khoa học máy tính\textsubscript{K50\ computer\ science course}, {?}}) and {({\textit{Normal}, \textit{List}, sinh viên\textsubscript{student}, có quê\textsubscript{has hometown}, Hà Nội\textsubscript{Hanoi}, {?}})}.% as presented in figure \ref{fig:kbqa}.

In the answer retrieval component, the ontology mapping module  maps the query tuples to ontology tuples: 
{({sinh viên\textsubscript{student}, học\textsubscript{enrolled}, lớp K50 khoa học máy tính\textsubscript{K50\ computer\ science\ course}})} and ({sinh viên\textsubscript{student}, có quê\textsubscript{has\ hometown}, Hà Nội\textsubscript{Hanoi}}).
For each ontology tuple, the answer extraction module finds all satisfied instances in the target ontology, and it then generates an answer based on the  \textit{And} question structure and the  \textit{List} question category. Figure \ref{fig:kbqa} shows the answer.

\subsection{Natural language question analysis component}

The natural language question analysis component  is the first component in any QA system.
When a question is asked, the task of this component is to convert the input question into an intermediate representation which is then used in the rest of the system.

KbQAS makes  use of the JAPE grammars in the
GATE framework \cite{CunninghamMBT02} to
specify semantic
annotation-based regular expression patterns for question analysis, in which
existing linguistic processing modules for Vietnamese
including word segmentation and part-of-speech
tagging \cite{PhamTP09} are wrapped as GATE plug-ins. The results of the wrapped plug-ins are annotations covering sentences  and segmented words. Each annotation has a set of feature-value pairs. For example, a word has a  \textit{category} feature storing its part-of-speech tag. This information can then be reused for further processing in subsequent modules. The new question analysis modules of preprocessing, syntactic analysis  and semantic analysis in KbQAS are specifically designed to handle Vietnamese questions using patterns over existing linguistic annotations.

\subsubsection{Preprocessing module}

\begin{figure}[h]
\centering
\includegraphics[width=7.5cm]{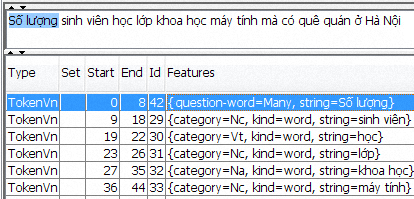}
\caption{Examples of \textit{TokenVn} annotations.}
\label{fig:tokenvn}
\end{figure}

The preprocessing module generates \textit{TokenVn} annotations representing a Vietnamese word with features, such as part-of-speech, as displayed in Figure \ref{fig:tokenvn}. Vietnamese is a monosyllabic language; hence, a word can contain more than one token. So there are words or word phrases which are indicative of the question categories, such as { ``phải không\textsubscript{is \ that / are \ there}''}, { ``là bao nhiêu\textsubscript{how\ many}''}, { ``ở đâu\textsubscript{where}''}, { ``khi nào\textsubscript{when}''} and { ``là cái gì\textsubscript{what}''}. However, the Vietnamese word segmentation module was not trained on the question domain. In this module, therefore, we identify those words or phrases and label them as single \textit{TokenVn} annotations with the  \textit{question-word} feature and its semantic category, like \textit{HowWhy\textsubscript{cause\ /\ method}}, \textit{YesNo\textsubscript{true\ or\ false}}, \textit{What\textsubscript{something}}, \textit{When\textsubscript{time\ /\ date}}, \textit{Where\textsubscript{location}},  \textit{Many\textsubscript{number}} or \textit{Who\textsubscript{person}}. In fact, this information will be used to create rules in the syntactic analysis module at a later stage.

We also label special words, such as abbreviations of  words on a special domain, and phrases that refer to a comparison, such as {``lớn hơn\textsubscript{greater\ than}''}, {``nhỏ hơn hoặc bằng\textsubscript{less\ than\ or\ equal\ to}''} and the like, by single \textit{TokenVn} annotations.

\begin{figure*}[ht]
\centering
\includegraphics[width=15cm]{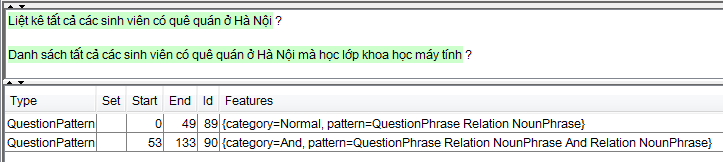}
\caption{Examples of question structure patterns.}
\label{fig:pattern}
\end{figure*}

\subsubsection{Syntactic analysis}
\label{ssubsec:sa}

The syntactic analysis module is responsible for identifying concepts, entities  and the relations between them in the input question. This module uses the \textit{TokenVn} annotations which are the output of the preprocessing module.

\begin{table}[h]
\caption{JAPE grammar for identifying Vietnamese noun phrases.}
\label{table1}
\begin{tabular}{|l|l|}
\hline
\textbf{(} \{TokenVn.category == ``Pn''\} \textbf{)}? & Quantity pronoun \\ 
%\hline
\textbf{(} \{TokenVn.category == ``Nu''\} |  & Concrete noun  \\
\ \ \{TokenVn.category == ``Nn''\} \textbf{)}? & Numeral noun  \\
%\hline
\textbf{(} \{TokenVn.string == ``cái''\} | & ``cái\textsubscript{the}'' \\
\ \ \{TokenVn.string == ``chiếc''\} \textbf{)}? & ``chiếc\textsubscript{the}'' \\
%\hline
\textbf{(} \{TokenVn.category == ``Nt''\} \textbf{)}? & Classifier noun \\
\textbf{(} \{TokenVn.category == ``Nc''\} | & Countable noun \\
\ \ \{TokenVn.category == ``Ng''\} | & Collective noun \\
\ \ \{TokenVn.category == ``Nu''\} | & \\
\ \ \{TokenVn.category == ``Na''\} | & Abstract noun \\
\ \ \{TokenVn.category == ``Np''\} \textbf{)}+ & Proper noun \\
%\hline
\textbf{(} \{TokenVn.category == ``Aa''\} | & Quality adjective \\
\ \ \{TokenVn.category == ``An''\} \textbf{)}? & Quantity adjective \\
%\hline
\textbf{(} \{TokenVn.string == ``này''\} | &  ``này\textsubscript{this;\ these}'' \\
\ \ \{TokenVn.string == ``kia''\} | & ``kia\textsubscript{that;\ those}'' \\
\ \ \{TokenVn.string == ``ấy''\} | & ``ấy\textsubscript{that;\ those}'' \\
\ \ \{TokenVn.string == ``đó''\} \textbf{)}? & ``đó\textsubscript{that;\ those}'' \\
\hline
\end{tabular}
\end{table}

Concepts and entities are normally expressed in noun phrases. Therefore, it is crucial to identify noun phrases in order to generate the {query tuple}. Based on the  Vietnamese language grammar \cite{DiepQB2005}, we use the {JAPE} grammars to specify patterns over annotations  as shown in Table \ref{table1}.
When a noun phrase is matched, a  \textit{NounPhrase} annotation is created to mark up the noun phrase. In addition, a \textit{type} feature of the \textit{NounPhrase} annotation is used to determine whether concept or entity is covered by the noun phrase, using the following heuristics:  if the noun phrase contains a single noun (not including numeral nouns) and does not contain a proper noun, it covers a {concept}. If the noun phrase contains a proper noun or  at least three single nouns, it covers an {entity}.  Otherwise, the \textit{type} feature value is determined by using a dictionary\footnote{The dictionary contains concepts which are extracted from the target ontology. However, there is no publicly available WordNet-like lexicon for Vietnamese. So we manually add synonyms of the extracted concepts to the dictionary.}.

Furthermore, the question phrases are detected by using the matched noun phrases and the question-words which are identified by the preprocessing module. \textit{QuestionPhrase} annotations are generated to cover the question phrases, with a  \textit{category} feature that gives  information about question categories. 

The next step is to identify {relation}s between noun phrases or between a noun phrase  and a question phrase. When a phrase is matched by one of the relation patterns, a  \textit{Relation} annotation is created to markup the relation. 
We use the following four grammar patterns to determine relation phrases:
%\\(Verb)+(NounPhrase$_{type==Concept}$)(Preposition)(Verb){?}
%\\(Verb)+((Preposition)(Verb){?}){?}
%\\((``có$_{have|has}$'')\textbf{|}(Verb))+(Adjective)(Preposition)(Verb){?} 
%\\(``có$_{have|has}$'')((NounPhrase$_{type==Concept}$)\textbf{|}(Adjective))(``là$_{is}$'')

\medskip

\begin{tabular}{l}
\hline
(Verb)+\\(Noun Phrase\textsubscript{type == Concept})\\(Preposition)(Verb){?} \\ 
\hline
(Verb)+((Preposition)(Verb){?}){?} \\
\hline
((``có\textsubscript{have/has}'')\textbf{|}(Verb))+\\(Adjective)\\(Preposition)\\(Verb){?} \\
\hline
(``có\textsubscript{have/has}'')\\((Noun Phrase\textsubscript{type == Concept})\textbf{|}(Adjective))\\(``là\textsubscript{is/are}'') \\
\hline
\end{tabular}

\medskip
For example, we can describe the first question {``Liệt kê tất cả các sinh viên có quê quán ở Hà Nội''} (``List all students whose  hometown is Hanoi'') in  Figure \ref{fig:pattern}, using  \textit{NounPhrase}, \textit{Relation} and \textit{QuestionPhrase}  annotations as follows:

\vspace{3pt}
[QuestionPhrase: Liệt kê\textsubscript{list} [NounPhrase: tất cả các sinh viên\textsubscript{all\ students}]] [Relation: có quê quán ở\textsubscript{have  hometown}] [NounPhrase: Hà Nội\textsubscript{Hanoi}]
\vspace{3pt}

The phrase {``có quê quán ở\textsubscript{have\ hometown}'' is the relation  linking the question phrase {``liệt kê tất cả các sinh viên\textsubscript{list\ all\ students}''} and the noun phrase {``Hà Nội\textsubscript{Hanoi}''}.

\subsubsection{Semantic analysis module}
\label{subsec:sa}

The semantic analysis module aims to identify the question structure and produce the query tuples  
{(sub-structure, question category, \emph{Term}$_1$, \emph{Relation}, \emph{Term}$_2$, \emph{Term}$_3$)} as the intermediate representation element of the input question, using the \textit{TokenVn}, \textit{NounPhrase}, \textit{Relation} and \textit{QuestionPhrase} annotations returned by the two previous modules.
Existing \textit{NounPhrase} annotations  and \textit{Relation}
annotations are potential candidates for terms and relations in the query tuples,
respectively. In addition, \textit{QuestionPhrase} annotations are used to detect the question category.

In the first KbQAS  version \cite{NguyenNP09}, following AquaLog \cite{LopezUMP07}, we developed an ad-hoc approach to detect structure patterns of questions and then use these patterns to generate  the intermediate representation elements. For example,  Figure \ref{fig:pattern} presents the detected structure patterns of the two example questions {``Liệt kê tất cả các sinh viên có quê quán ở Hà Nội''}  ({``List all students whose  hometown is Hanoi''}) and {``Danh sách tất cả các sinh viên có quê quán ở Hà  Nội mà học lớp khoa học máy tính''} ({``List all students enrolled in the computer science course, whose hometown is Hanoi''}). We can describe these questions by using annotations generated by the preprocessing and syntactic analysis modules  as follows:

\vspace{3pt}
[QuestionPhrase: Liệt kê tất cả các sinh viên\textsubscript{List all} \textsubscript{students}] [Relation: có quê quán ở\textsubscript{have\ hometown}] [NounPhrase: Hà Nội\textsubscript{Hanoi}]
\vspace{3pt}

and 

\vspace{3pt}

[QuestionPhrase: Liệt kê tất cả các sinh viên\textsubscript{List\ all} \textsubscript{students}] [Relation: có quê quán ở\textsubscript{have\ hometown}] [NounPhrase: Hà Nội\textsubscript{Hanoi}] [And:  [TokenVn: mà\textsubscript{and}]]  [Relation: học\textsubscript{enrolled}] [NounPhrase: lớp khoa học máy tính\textsubscript{computer\ science\ course}]

\vspace{3pt}

The intermediate representation element of an input question is created in a hard-wired manner linking every detected structure pattern via JAPE grammars. This hard-wired manner takes a lot of time and effort to handle new patterns. For example in Figure \ref{fig:pattern}, the hard-wired approach is unable to reuse the detected structure pattern of the first question to identify the structure pattern of the second question.
Since JAPE grammar rules were created in an ad-hoc manner, the hard-wired approach encounters  common difficulties in managing the interaction among rules and keeping consistency.

Consequently, in this module, we solve the mentioned difficulties by proposing a knowledge acquisition approach for the semantic analysis of input questions, as detailed in Section \ref{sec:rdrqa}.  In this paper, this is considered as the key innovation of KbQAS.

\subsection{Answer retrieval component}
\label{ssec:arc}

As presented in the first KbQAS  version \cite{NguyenNP09}, the answer retrieval component includes two  modules: ontology mapping and answer extraction, as shown in Figure \ref{Figure1}. It takes the intermediate representation  produced by the question analysis component and a target ontology as its input to generate an answer. To develop the answer retrieval component in KbQAS, we employed the relation similarity service component of  AquaLog  \cite{LopezUMP07}.

The task of the ontology mapping module is to map terms and relations in the query tuple to concepts, instances and relations in the target ontology by using string names. If an exact match is not possible, we use a string distance algorithm  \cite{VargasM04} and  the dictionary containing concepts and their synonyms to find near-matched elements from the target ontology, with the similarity measure above a certain threshold. 

In case  the ambiguity is still present, %, similar to  Aqualog \cite{LopezUMP07},  
 KbQAS  interacts with users by showing different options, and the users then choose the suitable ontology element. For example, given the question {``Liệt kê tất cả các sinh viên học lớp khoa học máy tính''} (``List all students enrolled in the computer science course''), the question analysis component produces a query tuple  ({\textit{Normal}, \textit{List}, sinh viên\textsubscript{student}, học\textsubscript{enrolled}, lớp khoa học máy tính\textsubscript{computer\ science\ course}, ?}). Because the ontology mapping module cannot find the exact instance corresponding to {``lớp khoa học máy tính \textsubscript{computer\ science\ course}''} in the target ontology, it requires the user to select between {``lớp K50 khoa học máy tính \textsubscript{K50\ computer\ science\ course}''} - an instance of class {``lớp\textsubscript{course}''}, and {``bộ môn khoa học máy tính\textsubscript{computer\ science\ department}''} - an instance of class {``bộ môn\textsubscript{department}''}.
 
 \begin{figure}[h]
\centering
\includegraphics[width=7.25cm]{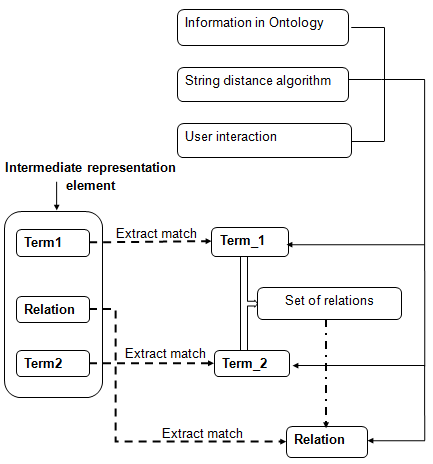}
\caption{Ontology mapping module for the query tuple with two terms and one relation.}
\label{fig:match}
\end{figure}

Following  AquaLog, for each query tuple, the result of the ontology mapping module is an ontology tuple where the terms and relations in the query tuple are now the corresponding elements from the target ontology.
How the ontology mapping module finds the corresponding elements from the target ontology depends on the question structure. For example, when the query tuple contains \emph{Term}$_1$, \emph{Term}$_2$ and \emph{Relation} with \emph{Term}$_3$ missing, the mapping process follows the diagram shown in Figure \ref{fig:match}. The mapping process first tries to match \emph{Term}$_1$ and \emph{Term}$_2$  with concepts or instances in the target ontology. Then the mapping process finds a set of potential relations between the two mapped concepts/instances from the target ontology. The ontology relation is finally identified by mapping  \emph{Relation} to a relation in the potential relation set, using a manner similar to mapping a term to a concept or an instance. 

With the ontology tuple, the answer extraction module finds all individuals of the  ontology concept corresponding to \emph{Term}$_1$, having the ontology relation with the ontology individual corresponding to \emph{Term}$_2$. The answer extraction module then returns the answer based on the question structure and question category. See  the definitions of question structure and question category types in Appendix \textbf{A} and Appendix \textbf{B}.

%%%%%%%%%%%%%%%%%%%%%%%%%%%%%%%%%%%%%%%%%%%%%%%%%%%%%%%%%%%%%%%%%%%%%
\section{Single Classification Ripple Down Rules for question analysis}
\label{sec:rdrqa}

As mentioned in Section \ref{subsec:sa}, due to the representation complexity  and the variety of question structures,  manually creating grammar rules in an ad-hoc manner is very expensive and error-prone. For example, such rule-based approaches as presented in \cite{LopezUMP07,NguyenNP09,PhanN10} manually defined a list of sequence pattern structures to analyze questions.
Since rules were created in an ad-hoc manner, these approaches share common difficulties in managing the interaction between rules and keeping consistency among them.

This section introduces our knowledge acquisition approach\footnote{The English  question analysis demonstration is available online at \url{http://150.65.242.39:8080/KbEnQA/}, and the Vietnamese question analysis demonstration is available online at \url{http://150.65.242.39:8080/KbVnQA/}.} to analyze natural language questions by applying the Single Classification Ripple Down Rules (SCRDR) methodology \cite{ComptonJ90,RichardsD09}  to acquire rules incrementally.  {Our contribution} focuses on the {semantic analysis module} by proposing a JAPE-like rule language and a systematic  processing to create rules in a manner that the interaction among rules is controlled  and  consistency is maintained. Compared to the first KbQAS  version \cite{NguyenNP09}, this is the key innovation of the current KbQAS version.

A SCRDR knowledge base is built to identify the question structures and to produce the query tuples as the intermediate representations of the input questions. We  outline the SCRDR methodology and propose a rule language for extracting the intermediate representation of a given  question in Section \ref{subsec:scrdr} and Section  \ref{subsec:rule}, respectively. We then illustrate the process of systematically constructing a SCRDR
knowledge base for analyzing questions in Section \ref{subsec:kap}. 

%\begin{figure}[ht]
%\centering
%\includegraphics[width=16.5cm]{fig_rdrpos.png}
%\caption{A part of SCRDR tree for English POS tagging \cite{NguyenNPP11}.}
%\label{fig:scrdrPOS}

%\end{figure}

\subsection{Single Classification Ripple Down Rules}
\label{subsec:scrdr}

This section presents the basic idea of  Single Classification Ripple Down Rules (SCRDR) \cite{ComptonJ90,RichardsD09} which inspired our knowledge acquisition approach for question analysis. A SCRDR tree  is a binary tree with two distinct types of edges.
These edges are typically called {\em except} and {\em false} edges.
Associated with each node in a tree is a {\em rule}.
A rule has the form: {\em if
} $\alpha$ {\em then} $\beta$ where $\alpha$ is called the {\em
condition} and $\beta$ is called  the {\em conclusion}.

 %Ripple Down Rules methodology allows one to add rules to a knowledge base incrementally without the need of a knowledge engineer. A new rule (i.e. an exception rule) is only created when the knowledge base performs unsatisfactorily on a given case. The rule represents an explanation for why the conclusion should be different from the knowledge base's conclusion on the case at hand.

\begin{figure}[h]
\centering
\includegraphics[width=6.125cm]{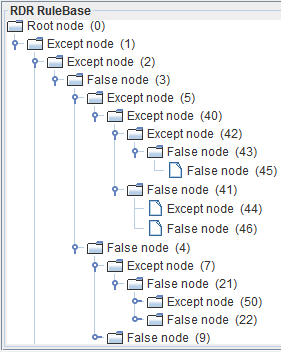}
\caption{A part of the SCRDR tree for English question analysis.}
\label{fig:enkb}
\end{figure}

\begin{figure*}[ht]
\centering
\includegraphics[width=15.75cm]{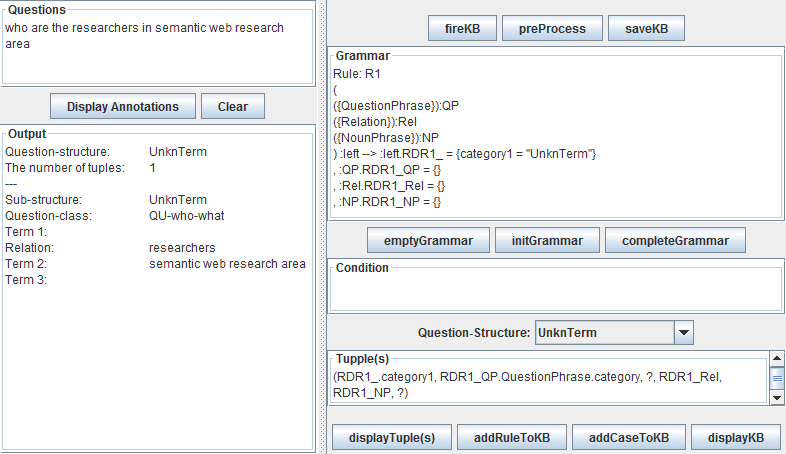}
\caption{The graphic user interface for knowledge base construction.}
\label{fig:gui}
\end{figure*}

Cases in SCRDR
are evaluated by passing a case to the root node of the SCRDR tree.  
At any node in the SCRDR tree, if the condition of the rule at a node $\eta$ is satisfied by the case (so the node $\eta$ \textit{fires}),
the case is passed on to the {\em except} child node of the node $\eta$ using the {\em except} edge if it exists; 
otherwise, the case is passed on to the {\em false} child node of the node $\eta$.
The conclusion given by this process is the conclusion from
the  node which {\em fired} last. 

Given the question ``Who are the partners involved in AKT project ?'' and the SCRDR tree in Figure \ref{fig:enkb}, it is satisfied by the  rule at the root node (0). Then it is passed to node (1) using the except edge. As the case satisfies the condition of the rule at node (1), it is passed to node (2) using the except edge. Because the case does not satisfy the condition of the rule at node (2), it is then passed to  node (3) using the false edge. As the case satisfies  the conditions of the rules at nodes (3), (5) and (40), it is passed to node (42), using except edges. Since the case does not satisfy the conditions of the rules at nodes (42), (43) and (45), we have the evaluation path (0)-(1)-(2)-(3)-(5)-(40)-(42)-(43)-(45) with the last fired node (40). Given another   case of ``In which projects is enrico motta working on'', it satisfies the conditions of the rules at nodes (0), (1) and (2); as node (2) has no except child node, we have the evaluation path (0)-(1)-(2) and the last fired node (2).
 
A new node containing a new exception rule is added to an SCRDR tree when the evaluation process returns an \textit{incorrect} conclusion.  
The new exception node is attached to the last node in the evaluation path of the given case as an  {\em except}
edge if the last node is the {fired} node; otherwise, it is attached as an  {\em false} edge. 
 
 To ensure that a conclusion is always reached, the root node, called the {\em default} node, typically contains a trivial condition which is always satisfied.  
 The rule at the default node, the default rule, is the unique rule which is not an exception rule of any other rule. For example, the default rule ``\textit{if} \textbf{True} \textit{then} \textbf{null}'' from the SCRDR tree in Figure \ref{fig:enkb} means that its \textit{True} condition satisfies every question, however, its  \textit{null} conclusion produces an empty intermediate representation element for every question. Started with a  SCRDR knowledge base consisting of only the default node, the process of building the knowledge base can be performed automatically \cite{NguyenNPP11,nguyen-EtAl:2014:Demos}  or manually \cite{PhamH06,NguyenNP11}.% Section \ref{subsec:kap} will demonstrate the manual construction process of the SCRDR tree displayed in Figure \ref{fig:enkb}.

 In  the SCRDR tree from Figure \ref{fig:enkb},  the rule at node (1) (simply, rule 1) is an exception rule of the default rule 0. Rule 2 is an exception rule of rule 1. 
As node (3) is the false-child node of node (2), the rule 3 is also an exception rule of rule 1. Furthermore, both rules 4 and 9 are also exception rules of  rule 1. Similarly, all rules 40, 41 and 46 are exception rules of  rule 5 while all rules 42, 43 and 45 are  exception rules of  rule 40. Therefore, the exception structure of the SCRDR tree
extends to 5 levels, for examples: rules 1 at layer 1; rules
2, 3, 4 and 9 at layer 2; rules 5, 7, 21 and
22 at layer 3; and rules 40, 41, 46 and 50 at layer 4; and rules 42, 43, 44 and 45 at the layer 5  exception structure.

\subsection{Rule language}
\label{subsec:rule}

A rule is composed of a condition part and a conclusion part. 
A condition is a regular expression pattern over annotations using JAPE grammars in GATE \cite{CunninghamMBT02}.
It can also post new annotations over matched phrases of the pattern's sub-components.
As annotations have feature-value pairs, we can impose constraints on the annotations in the pattern by specifying that a feature of an annotation must have a particular value. The following example shows the posting of an annotation over the matched phrase:

\vspace{3pt}
{\footnotesize {\ttfamily 
\noindent ((\{TokenVn.string == ``liệt kê\textsubscript{list}''\} |  \\
\{TokenVn.string == ``chỉ ra\textsubscript{show}''\}) \\
\{NounPhrase.type == ``Concept''\}):qp \\
$ \dashrightarrow $ :qp.QuestionPhrase = \{category = ``List''\}
}}
\vspace{3pt}

 Every complete pattern  followed by a label must be enclosed by round brackets. In the above pattern, the label is \texttt{qp}. The pattern would match phrases starting with a \textit{ TokenVn} annotation covering either the word { ``liệt kê\textsubscript{list}''} or the word { ``chỉ ra\textsubscript{show}''}, followed by a \textit{ NounPhrase} annotation covering a \textit{concept}-typed noun phrase. 
When applying this pattern on a text fragment,  \textit{QuestionPhrase} annotations having the \textit{category} feature with its \textit{List} value would be posted over phrases matched by the pattern. 
Furthermore, the condition part of a rule can include additional constraints. See  examples  of the additional constraints from the constructions of rules (40) and (45) in Section \ref{subsec:kap}. 

The conclusion part of a rule produces an intermediate representation containing the question structure and the query tuples, where each attribute
in the query tuples is specified by a newly posted annotation  from matching the rule's condition, in the following order:

\vspace{3pt}
{(sub-structure, question category, \emph{Term}$_1$, \emph{Relation}, \emph{Term}$_2$, \emph{Term}$_3$)}
\vspace{3pt}

All newly posted annotations have the same  \textit{RDR} prefix and the rule index so that a rule can refer to annotations of its parent rules.
Examples of rules and how rules are created and stored in an exception structure will be explained in details in Section \ref{subsec:kap}.

Given an input question,  the condition of a rule is satisfied if the whole input question is matched by the condition pattern. 
The conclusion of the fired rule produces the intermediate representation element of the input question. 
To create rules for matching the structures of questions, we use patterns over annotations %such as TokenVn, NounPhrase, Relation, annotations capturing question-phrases like QUTerm, QU-E-L-MC (Entity, List, ManyClass)\ldots and their features which are
 returned by the preprocessing and syntactic analysis modules.

\subsection{Knowledge acquisition process}
\label{subsec:kap}

 Our approach is
language-independent, because the main focus is on the process of creating
the rule-based system. 
  The language-specific part is in the rules itself. So, in this section, we illustrate the process of building a SCRDR knowledge base to analyze English questions.  Figure \ref{fig:gui} shows the graphic user interface to construct SCRDR knowledge bases.

We  reused the JAPE grammars which were developed to identify  noun phrases, question phrases and relation phrases in   AquaLog \cite{LopezUMP07}. Based on  \textit{Token} annotations which are generated as output of the English tokenizer, sentence splitter and part-of-speech tagger in the GATE framework \cite{CunninghamMBT02}, the JAPE grammars produce  \textit{NounPhrase}\footnote{Here annotations are generated without any concept or entity type information.}, \textit{QuestionPhrase} and \textit{Relation} annotations,  and  other annotation kinds such as \textit{Noun}, \textit{Verb} or \textit{Preps} annotations for covering  nouns,  verbs or  prepositions, respectively. We also reused question category definitions from AquaLog. 

For  illustrations in Section \ref{sss:re} and Section \ref{sssection:qam}, we employed a training set of 170 English questions\footnote{\url{http://technologies.kmi.open.ac.uk/aqualog/examples.html}}, which AquaLog \cite{LopezUMP07} analyzed successfully, to construct the SCRDR knowledge base  in Figure \ref{fig:enkb}. These questions concern the Knowledge Media Institute and its research area on the semantic web. 
%The rest of this section describes how the knowledge base building process works. 

\subsubsection{Reusing detected question structures}
\label{sss:re}

In contrast to the example in Section \ref{subsec:sa} with respect to Figure \ref{fig:pattern}, we start with demonstrations of reusing detected question structure patterns.

\vspace{-3pt}

\begin{figure}[ht]
\centering
\includegraphics[width=7.125cm]{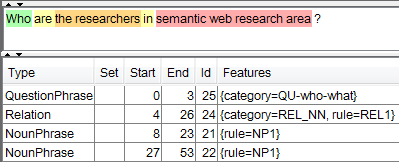}
\caption{Examples of annotations.}
\label{fig:annotations1}
\end{figure}

\vspace{-3pt}

For the question  ``Who are the researchers in semantic web research area ?'', we can represent this question using \textit{NounPhrase}, \textit{QuestionPhrase} and \textit{Relation} annotations as shown in Figure  \ref{fig:annotations1} as follows:

\vspace{3pt}

[QuestionPhrase: Who] [Relation: are the researchers in] [NounPhrase: semantic web research area]
\vspace{3pt}

Supposed we start with a knowledge base containing only the default rule \textbf{R0}. Given the question,  \textbf{R0} is the fired rule that gives an incorrect conclusion of an empty intermediate representation element. This can be corrected by adding the following rule \textbf{R1} as an exception rule of  \textbf{R0}. In the knowledge base, node (1) containing  \textbf{R1} is added as the except-child node of the default node, as shown in Figure \ref{fig:enkb}.  

\vspace{5pt}

\noindent \textbf{Rule: R1}

{\small{\ttfamily

\noindent (

\noindent (\{QuestionPhrase\}):qp

\noindent (\{Relation\}):rel

\noindent (\{NounPhrase\}):np

\noindent ):left 

\noindent  $ \dashrightarrow $ :left.RDR1\_  = \{category1 = ``UnknTerm''\} 

\noindent , :qp.RDR1\_QP = \{\} 

\noindent , :rel.RDR1\_Rel = \{\} 

\noindent , :np.RDR1\_NP = \{\} 

\vspace{3pt}
}}
\noindent \textbf{Conclusion:}

\noindent  \textit{UnknTerm} question structure and one query tuple {\small ({RDR1\_.category1, RDR1\_QP.QuestionPhrase.category,  ?, RDR1\_Rel, RDR1\_NP, ?})}
\vspace{5pt}

If the condition of  {\textbf{R1}} matches the whole input question, a new \textit{RDR1\_} annotation will be created to entirely cover the  input question. In addition, new annotations  \textit{RDR1\_QP},\textit{ RDR1\_Rel} and \textit{RDR1\_NP} will be created to cover the same question phrase, relation phrase and noun phrase as the \textit{QuestionPhrase}, \textit{Relation} and \textit{NounPhrase} annotations, respectively.

 When node (1) fired, the input question has one query tuple where the {sub-structure} attribute takes the value of the   \textit{category1} feature of the \textit{RDR1\_} annotation; the {question category} attribute takes the value of the \textit {category} feature of the \textit{QuestionPhrase} annotation which is in the same span as the \textit{RDR1\_QP} annotation. In addition, the \emph{Relation} and \emph{Term}$_2$ attributes take values of the strings covered by the \textit{RDR1\_Rel} and \textit{RDR1\_NP} annotations, respectively,  while {\emph{Term}$_1$} and {\emph{Term}$_3$}  are missing. The example of firing the question at node (1) is displayed in Figure \ref{fig:gui}.

%\medskip
%When it came to the question: 

%\textit{in which projects is Enrico Motta working on?}

%[Preposition: in] [RDR1\_: [RDR1\_QP: which projects] [RDR1\_Rel: is] [RDR1\_NP: Enrico Motta]]     [Relation: working on]

%The fired rule is rule \textbf{R1}, however, it produces an incorrect conclusion of question structure \textit{UnknTerm}  and query tuple \textit{(UnknTerm, QU-whichClass, ?, ?, Enrico Motta)}  as the RDR1\_ annotation only covers a part of the question and \textit{``is''} is not considered as a relation. The below rule \textbf{R2} would be appended as an exception rule of the rule \textbf{R1} to return a correct intermediate representation of the question:
 
\vspace{3pt}
Assume that, in addition to \textbf{R0} and \textbf{R1}, the current knowledge base contains rule \textbf{R2} as an exception rule of  \textbf{R1},  for which node (2) containing  \textbf{R2} is the except-child node of node (1), as shown in Figure \ref{fig:enkb}.  

\vspace{3pt}
For the question
 {``Which universities are Knowledge Media Institute collaborating with ?''}, the following annotation-based representation is constructed:

\vspace{3pt}
{\small [RDR1\_: [RDR1\_QP: Which universities] [RDR1\_Rel: are] [RDR1\_NP: Knowledge Media Institute]] [Relation: collaborating with]}
\vspace{3pt}

We have the evaluation path of (0)-(1)-(2) with the last fired node (1).  However,  \textbf{R1} produces an incorrect conclusion of the \textit{UnknTerm} question structure  and one query tuple ({\textit{UnknTerm}, \textit{QU-whichClass}, ?, ?, Knowledge Media Institute, ?}). It is because the \textit{RDR1\_} annotation only covers a part of the question and {``are''} is not considered as a relation. The following rule \textbf{R3} is added as an exception rule of  \textbf{R1}:

\vspace{5pt}

\noindent \textbf{Rule: R3}

{\small{\ttfamily

\noindent (

\noindent \{RDR1\_\} 
 (\{Relation\}):rel

\noindent ):left 

\noindent $ \dashrightarrow $ :left.RDR3\_  = \{category1 = ``Normal''\} 

\noindent , :rel.RDR3\_Rel = \{\} 

\vspace{3pt}

}} 

\noindent \textbf{Conclusion:}  

\noindent \textit{Normal} question structure and one query tuple 
\\  {\small ({RDR3\_.category1,  RDR1\_QP.QuestionPhrase.category, \\ RDR1\_QP, RDR3\_Rel, RDR1\_NP, ?})}

\vspace{5pt}

In the knowledge base, node (3) containing  \textbf{R3} is appended as the false-child node of node (2) which is the last node in the evaluation path. Regarding  the input question ``{Which universities are Knowledge Media Institute collaborating with ?}'', we have a new evaluation path of (0)-(1)-(2)-(3) with the last fired node (3). So  \textbf{R3} produces a correct intermediate representation element of the question, consisting of the \textit{Normal} question structure  and one query tuple ({\textit{Normal}, \textit{QU-whichClass}, universities, collaborating, Knowledge Media Institute, ?}).

\vspace{3pt}
Subsequently,   another question makes an addition of  rule \textbf{R4} which is also an exception rule of   \textbf{R1}. In the knowledge base, node (4) containing  \textbf{R4} is appended as the false-child node of  node (3).
\vspace{3pt} 

For the question  {``Who are the partners involved in AKT project ?''}, we have an annotation-based representation as follows:

\vspace{3pt}
{\small [RDR3\_: [RDR1\_QP: Who] [RDR1\_Rel: are] [RDR1\_NP: the partners] [RDR3\_Rel: involved in]] [NounPhrase: AKT project]}
\vspace{3pt}

We have the evaluation path (0)-(1)-(2)-(3) and node (3) is the last fired node. But  \textbf{R3} returns a wrong conclusion as the \textit{RDR3\_} annotation covers a part of the question. The following rule \textbf{R5} is added as an exception rule of  \textbf{R3}  to correct the returned conclusion:

\vspace{5pt}

\noindent \textbf{Rule: R5}

{\small{\ttfamily

\noindent (

\noindent \{RDR3\_\} 
(\{NounPhrase\}):np

\noindent ):left 

\noindent $ \dashrightarrow $ :left.RDR5\_  = \{category1 = ``Normal''\} 

\noindent , :np.RDR5\_NP = \{\} 

}}

\vspace{3pt}
\noindent \textbf{Conclusion:} 

\noindent  \textit{Normal} question structure  and one query tuple \\ {\small (RDR5\_.category1, { RDR1\_QP.QuestionPhrase.category}, RDR1\_NP, RDR3\_Rel, RDR5\_NP, ?)}

\vspace{5pt}

As node (3) is the last node in the evaluation path,  node (5) containing  \textbf{R5} is attached as the except-child node of node (3). Using  \textbf{R5}, we get a correct conclusion consisting of  the \textit{Normal} question structure  and one query tuple ({\textit{Normal}, \textit{QU-who-what}, partners, involved, AKT project, ?}).

\vspace{3pt}

\subsubsection{Solving question structure ambiguities}
\label{sssection:qam}

\vspace{3pt}

 The process of adding the rules above illustrates the ability of quickly handling new question structure patterns of our knowledge acquisition approach against the ad-hoc approaches \cite{LopezUMP07,NguyenNP09}.   The following examples demonstrate the ability of our approach to solve question structure ambiguities.   

\vspace{3pt}
For the question {``Which researchers wrote publications related to semantic portals ?''}, the following
representation is produced:

\vspace{3pt}
{\small [RDR5\_: [RDR1\_QP: Which researchers] [RDR1\_Rel: wrote] [RDR1\_NP: publications] [RDR3\_Rel: related to] [RDR5\_NP: semantic portals]]}
\vspace{3pt}

This question is fired at node (5) which is the last node in the evaluation path (0)-(1)-(2)-(3)-(5). But  \textbf{R5} gives a wrong conclusion of the \textit{Normal} question structure  and one query tuple ({\textit{Normal}, \textit{QU-whichClass}, publications,  related to, semantic portals, ?}). We add the following rule  \textbf{R40} as an exception rule of  \textbf{R5} to correct the conclusion returned by \textbf{R5}:

\vspace{5pt}

\noindent \textbf{Rule: R40}

{\small{\ttfamily

\noindent (

\noindent \{RDR5\_\}

\noindent ):left 

\noindent $ \dashrightarrow $ :left.RDR40\_  = \{category1 = ``Normal'', category2 = ``Normal''\} 
}}
\vspace{3pt}

\noindent \textbf{Condition:} 

{\small{\ttfamily
\noindent {\footnotesize RDR1\_QP.hasAnno == QuestionPhrase.category == QU-whichClass}

\vspace{3pt}

 }}

\noindent \textbf{Conclusion:}  

\noindent  \textit{Clause} question structure\footnote{A \textit{Clause} structure question has two query tuples   where the answer returned for the second query tuple indicates the missing \emph{Term}$_2$ attribute in the first query tuple. See more details of our question structure definitions in  appendix A.}
and two query tuples \\ {\small ({RDR40\_.category1, RDR1\_QP.QuestionPhrase.category, RDR1\_QP, RDR1\_Rel, ?, ?}) }
and \\ {\small ({RDR40\_.category2, RDR1\_QP.QuestionPhrase.category, RDR1\_NP, RDR3\_Rel, RDR5\_NP, ?})} 

\vspace{5pt}

 The extra annotation constraint of \textit{hasAnno} requires that the text covered by an annotation must contain another specified annotation. For example, the additional condition in  \textbf{R40}  only matches the  \textit{RDR1\_QP}  annotation that has a \textit{QuestionPhrase} annotation covering its substring\footnote{A whole string is also considered as its substring.}. Additionally, this  \textit{QuestionPhrase} annotation must has ``{QU-whichClass}'' as the value of  its \textit{category} feature.
 
In the knowledge base, node (40) containing \textbf{R40} is added as the except-child node of node (5). Given the question, the last fired node now is node (40); and the conclusion of \textbf{R40} produces a correct intermediate representation  consisting  of the  \textit{Clause} question structure and two query tuples ({\textit{Normal}, \textit{QU-whichClass}, researchers, wrote, ?, ?}) and ({\textit{Normal}, \textit{QU-whichClass}, publications, related to, semantic portals, ?}).

\vspace{3pt}
For the question 
{``Which projects sponsored by eprsc are related to semantic web ?''}, we have  part-of-speech and annotation-based representations  as follows:

Which/WDT projects/NNS sponsored/VBN by/IN eprsc/NN are/VBP related/VBN to/TO semantic/JJ web/NN 

\vspace{3pt}

 [RDR40\_: [RDR1\_QP: [QuestionPhrase \textsubscript{category  = } \textsubscript{QU-whichClass}: Which projects]] [RDR1\_Rel:  sponsored by] [RDR1\_NP: eprsc] [RDR3\_Rel: are related to] [RDR5\_NP: semantic web]]

\vspace{3pt}

The current knowledge base generates an evaluation path (0)-(1)-(2)-(3)-(5)-(40)-(42)-(43) with the last fired node (40). However, \textbf{R40} returns a wrong conclusion with the \textit{Clause} question structure  and two query tuples ({\textit{Normal}, \textit{QU-whichClass}, projects, sponsored, ?, ?}) and ({\textit{Normal}, \textit{QU-whichClass}, eprsc, related to, semantic web, ?}) since \emph{Term}$_1$ cannot be assigned to the instance ``eprsc''. The following  rule \textbf{R45} which is a new exception rule of \textbf{R40} is added to correct the conclusion given by  \textbf{R40}:

\vspace{5pt}

\noindent \textbf{Rule: R45}

{\small{\ttfamily

\noindent (

\noindent \{RDR40\_\}

\noindent ):left 

\noindent $ \dashrightarrow $ :left.RDR45\_  = \{category1 = ``Normal'', category2 = ``Normal''\} 

\vspace{3pt}
}}
\noindent \textbf{Condition}: 

 {\small{\ttfamily

\noindent RDR1\_Rel.hasAnno == Token.category == VBN

\vspace{3pt}

}}

\noindent \textbf{Conclusion}:  

\noindent \textit{And}  question structure and two query tuples  \\ {\small ({RDR45\_.category1, RDR1\_QP.QuestionPhrase.category, RDR1\_QP, RDR1\_Rel, RDR1\_NP, ?})} and \\ 
{\small({RDR45\_.category2, RDR1\_QP.QuestionPhrase.category, RDR1\_QP, RDR3\_Rel, RDR5\_NP, ?})}
\vspace{5pt}

 \textbf{R45} enables to return a correct intermediate representation element for the question with the \textit{And} question structure  and  two query tuples ({\textit{Normal}, \textit{QU-whichClass}, projects, sponsored, eprsc, ?}) and ({\textit{Normal}, \textit{QU-whichClass}, projects, related to, semantic web, ?}). In  the knowledge base, the associated node (45) is attached as the false-child node of  node (43). 
 
\vspace{5pt}
 
 \subsubsection{Porting to other domains} 
 
\vspace{5pt}
 
As illustrated in Section \ref{sss:re} and Section \ref{sssection:qam}, using the set of 170 questions from AquaLog \cite{LopezUMP07}, we constructed a knowledge base of 59 rules for question analysis. Similarly, in this section, we illustrate the process of adding more exception rules into the knowledge base to handle DBpedia and biomedical test questions. 

\vspace{3pt}
For the DBpedia test question ``{Which presidents of the United States had more than three children ?}'', the following representations are constructed:

\vspace{3pt}

{\small Which/WDT presidents/NNS of/IN the/DT United/NNP States/NNPS had/VBD more/JJR than/IN three/CD children/NNS}

\vspace{3pt}

{\small[RDR27\_:  [RDR10\_: [RDR10\_QP:  Which presidents] [Preps: of] [RDR10\_NP: the United States]]    [RDR27\_Rel: had more than] [RDR27\_NP: three children]]}

\vspace{3pt}

The last fired node for this DBpedia question is node (27). However, the conclusion of  rule \textbf{R27} at node (27) produced an incorrect intermediate representation element %of the ``ThreeTerm'' question structure and a query tuple \textit{(ThreeTerm, QU-whichClass, presidents, had more than, three children, United States)} 
 for the question. So a new exception rule  of \textbf{R27} is added to the knowledge base to correct the conclusion returned by \textbf{R27} as follows:

\vspace{5pt}

\noindent \textbf{Rule: R67}

{\small{\ttfamily

\noindent (

\noindent \{RDR10\_\} \{Verb\}

\noindent (\{Token.category == JJR\} 

\noindent \{Token.string == than\} 

\noindent \{Token.category == CD\}):cp

\noindent (\{Noun\}):np

\noindent ):left 

\noindent $ \dashrightarrow $ :left.RDR67\_  =  \{category1 = ``Compare'', category2 = ``UnknRel''\} 

\noindent , :cp.RDR67\_Compare = \{\}

\noindent , :np.RDR67\_NP = \{\}

\vspace{3pt}

}}

\noindent \textbf{Conclusion:}   

\noindent  \textit{Clause} question structure and two query tuples  \\ {\small ({RDR67\_.category1, RDR10\_QP.QuestionPhrase.category, ? , RDR67\_NP, ?, RDR67\_Compare})} and \\ 
{\small({RDR67\_.category2, RDR10\_QP.QuestionPhrase.category, RDR10\_QP, ?, RDR10\_NP, ?})}

\vspace{5pt}

Given the question, \textbf{R67} produces a correct intermediate representation element of  the \textit{Clause} question structure and two query tuples  ({\textit{Compare}, \textit{QU-whichClass}, ?, children, ?, more than three}) and ({\textit{UnknRel}, \textit{QU-whichClass}, presidents, ?, United States, ?}).

\vspace{3pt}
 
For the biomedical test question  ``{List drugs that lead to strokes and arthrosis}'', we have the following representations:

\vspace{3pt}

 List/NN drugs/NNS that/WDT lead/VBP to/TO strokes/NNS and/CC arthrosis/NNS

\vspace{3pt}

[QuestionPhrase: List drugs] [RDR1\_: [RDR1\_QP: that] [RDR1\_Rel: lead to] [RDR1\_NP: strokes and arthrosis]]

\vspace{3pt}

The last fired node  for this biomedical question is node (1). However, \textbf{R1} returned an incorrect intermediate representation element.  So a new exception rule of \textbf{R1} is added to the knowledge base as follows:

\vspace{5pt}

\noindent \textbf{Rule: R80}

{\small{\ttfamily

\noindent (

\noindent (\{QuestionPhrase\}):qp

\noindent \{RDR1\_QP\} \{RDR1\_Rel\}

\noindent (\{Noun\}):np1

\noindent \{Token.category == CC\}

\noindent (\{Noun\}):np2

\noindent ):left 

\noindent  $ \dashrightarrow $  :left.RDR80\_  = \{category1 = ``Normal'', category2 = ``Normal''\} 

\noindent , :qp.RDR80\_QP = \{\}

\noindent , :np1.RDR80\_NP1 = \{\}

\noindent , :np2.RDR80\_NP2 = \{\}

\vspace{3pt}

}}

\noindent \textbf{Condition}: 

{\small{\ttfamily
\noindent   RDR80\_QP.hasAnno   == Noun
}}

\vspace{3pt}

\noindent \textbf{Conclusion}:  

\noindent \textit{And}  question structure and two query tuples  \\ {\small ({RDR80\_.category1, RDR80\_QP.QuestionPhrase.category, RDR80\_QP, RDR1\_Rel, RDR80\_NP1, ?})} and \\ 
{\small({RDR80\_.category2, RDR80\_QP.QuestionPhrase.category, RDR80\_QP, RDR1\_Rel, RDR80\_NP2, ?})}

\vspace{5pt}

Given the question, \textbf{R80} returns  a correct intermediate representation element of  the \textit{And}  question structure and two query tuples ({\textit{Normal}, \textit{QU-listClass}, drugs, lead to, strokes, ?}) and ({\textit{Normal}, \textit{QU-listClass}, drugs, lead to, arthrosis, ?}).
  
\section{Experiments}
\label{sec:experiments}

In KbQAS, the question analysis component employs our language-independent knowledge acquisition approach, while the  answer retrieval component produces answers from a domain-specific Vietnamese ontology.  So we separately evaluate the question analysis and answer retrieval components in Section \ref{subsec:eaq} and Section \ref{subsec:kbqas}, respectively. 

\subsection{Experiments on analyzing questions}
\label{subsec:eaq}

This section indicates the abilities of our question analysis approach for quickly building a new knowledge base and easily adapting to a new domain and a new language. We evaluate both our approaches of ad-hoc manner (see Section \ref{subsec:sa}) and knowledge acquisition (see Section \ref{sec:rdrqa}) on Vietnamese question analysis, and then present the experiment of building a  knowledge base for processing English questions.

\subsubsection{Question analysis for Vietnamese}
\label{sec:qav}

We used a training set of 400 questions of various
structures generated by four volunteer students. We then evaluated our question analysis approach on an unseen list of 88 questions related to the VNU University of Engineering and Technology, Vietnam. 
In this experiment, we also compare both our ad-hoc and knowledge acquisition approaches   for question analysis, using the same training set of 400 questions and test set of 88 questions. 

\begin{table}[ht]
\caption{Time to create rules and number of successfully analyzed questions.}
\label{tab:correctResult}
\begin{tabular}{l l l}
\hline Type & Time & \#questions\\
\hline 
	Ad-hoc & 75 hours & 70/88 (79.5\%) \\
	Knowledge acquisition  & 5 hours & 74/88 (\textbf{84.1\%}) \\
\hline
\end{tabular}
\end{table}

With our first approach it took about 75 hours  to create rules in an ad-hoc manner, as shown in Table \ref{tab:correctResult}. In contrast, with our second approach it took 13 hours to build a Vietnamese knowledge base of rules for question analysis. However, most of the time was spent  looking at questions to determine the question structures and the phrases which would be extracted to create  intermediate representation elements. So the actual time to create rules in the knowledge base was about 5 hours in total.

\begin{table}[ht]
\caption{Number of  exception rules in each layer in our Vietnamese knowledge base for question analysis.}
\label{tab:exceptRule}
\begin{tabular}{l@{\quad} l}
\hline Layer & Number of rules \\
\hline 1 & 26 \\
		2 & 41 \\
		3 & 20 \\
		4 & 4 \\
\hline
%???
\end{tabular}
\end{table}

The knowledge base consists of the default rule and 91 exception rules. 
Table \ref{tab:exceptRule} details the number of exception rules in each layer where every rule in layer $n$ is an exception rule of a rule in layer $n-1$. The only rule which is not an exception rule of any rule  is the default rule at layer 0. This indicates that the exception structure is indeed present and even extends to 4 levels.

%While table \ref{tab:numRuleV} gives the number of rules corresponding with each question structure type in the built knowledge based for Vietnamese. In our built knowledge base, there are 9 rules containing \textit{condition} constraints.

%This indicates that the cognitive load to create rules in the second approach is much less compared to that in the first one as in our case, we do not have to consider other rules when crafting a new rule.

Table \ref{tab:correctResult} also shows the number of successfully analyzed questions for each approach. By using the  knowledge base to resolve ambiguous cases, our knowledge acquisition approach  performs  better than our ad-hoc approach. 
Furthermore, Table \ref{tab:errorResult} provides the error sources for our knowledge acquisition approach, in which most errors come from unexpected question structure patterns.
This can be rectified by adding more exception rules to the current knowledge base, especially when having a large training set that contains a variety of  question structure patterns.

\begin{table}[ht]
\caption{Number of incorrectly analyzed questions accounted for the knowledge acquisition approach.}
\label{tab:errorResult}
\begin{tabular}{p{5.5cm}@{\quad} l}
\hline Reason & \#questions\\
\hline Unknown structure patterns & 12/88\\
	   Word segmentation and part-of-speech tagging modules were not trained on question domain & 2/88\\
\hline
\end{tabular}
\end{table}

For another example, our knowledge acquisition approach did not return a correct intermediate representation element for the question {``Vũ Tiến Thành có quê và có mã sinh viên là gì ?''}  ({``What is the hometown and student code of Vu Tien Thanh ?''}) because the existing linguistic processing modules for Vietnamese \cite{PhamTP09}, including word segmentation and part-of-speech tagging, were not trained on the question domain. So these two modules assign the word {``quê\textsubscript{hometown}''} as an adjective  instead of a noun. Thus,  {``quê\textsubscript{hometown}''} is not covered by  a \textit{NounPhrase} annotation, leading to an unrecognized structure pattern.

\begin{table}[ht]
\caption{Number of  rules  in the question analysis knowledge bases for Vietnamese (\#RV) and English (\#RE); number of Vietnamese test questions (\#TQ) and number of Vietnamese correctly answered questions (\#CA) corresponding to each question structure type (QST).}
\label{tab:numRuleVE}
\begin{tabular}{|l|l|l|l|l|}
\hline QST & \#RV & \#CA & \#TQ & \#RE\\
\hline 
		Definition & 2 & 1 & 2/2 & 4 \\
		UnknRel & 4 & 4 & 4/7 & 6 \\
		UnknTerm & 7 & 6 & 7/7 & 4 \\
		Normal & 7 & 7 & 7/7 & 11 \\
		Affirm & 10 & 5 & 5/5 & 5\\
		Compare & 5 &  0 & 2/4 & 8 \\
		ThreeTerm & 9 & 7 & 7/10 &  6 \\
		Affirm\_3Term & 5 & 4 & 4/4 & 2\\		
		And & 9 & 7 &  8/8 & 21 \\
		Or & 23 & 18 & 21/24 & 1 \\
		Affirm\_MoreTuples & 3 & 1 &  2/3 &  1\\
		Clause & 6 & 0 & 4/5 & 20\\
		Combine & 1 & 1 & 1/2 & 0\\
\hline
Total & 91 &61 &  74/88 & 89 \\
\hline
\end{tabular}
\end{table}

Regarding  a question structure-based evaluation, Table \ref{tab:numRuleVE} presents the number of rules in the Vietnamese knowledge base and  number of test questions,  corresponding to each question structure type. For example, the cell at the second row and  the fourth column of Table \ref{tab:numRuleVE} means that, in 7 test questions tending to have the  \textit{UnknRel} question structure, there are 4 test questions correctly analyzed.

\subsubsection{Question analysis for English}
\label{sec:qae}

For the experiment in English, we firstly used a set of 170 English questions\footnote{\url{http://technologies.kmi.open.ac.uk/aqualog/examples.html}}, which AquaLog \cite{LopezUMP07} analyzed successfully. These questions are about the Knowledge Media Institute and its research area on the semantic web.  Using this question set, we constructed a knowledge base of  59 rules  for question analysis. It took 7 hours to build the knowledge base, including 3 hours of actual time to create all rules. 
We then evaluated the knowledge base using a set of 50 DBpedia test questions  from the QALD-1 workshop and another set of 25 biomedical test questions from the QALD-4 workshop.\footnote{\url{http://www.sc.cit-ec.uni-bielefeld.de/qald/}}

\begin{table}[ht]
\caption{Test results of the knowledge base of 59 rules for question analysis on DBpedia and biomedical domains.}
\label{tab:kbEvalResult}
\begin{tabular}{p{3.75cm}@{\quad} | l | l }
\hline Factor & DBpedia & Biomedical\\
\hline
Successfully processed & 24/50 & 9/25 \\
\hline 
Unknown structure patterns & 18/50 & 9/25 \\
 
Incorrect word segmentation & 3/50 & 3/25 \\

Incorrect part-of-speech tagging & 5/50 & 4/25  \\
\hline
\end{tabular}
\end{table}

Table \ref{tab:kbEvalResult} presents evaluation results of analyzing the test questions from the DBpedia and biomedical domains, using the knowledge base of 59 rules for question analysis. It is not surprising that most errors come from unknown question structure patterns. Furthermore, just as in Vietnamese, the existing linguistic processing modules in the GATE framework \cite{CunninghamMBT02}, including the English tokenizer  and part-of-speech tagger, are also  error sources, leading to unrecognized structure patterns. For example, such questions  as ``{Which U.S. states possess gold minerals ?}'' and ``{Which drugs have a water solubility of 2.78e-01mg/mL ?}'' are tokenized into ``{Which U . S . states possess gold minerals ?}'' and ``{Which drugs have a water solubility of 2 . 78 e- 01 mg / mL ?}'', respectively. In addition, such other questions as ``{Which river does the Brooklyn Bridge cross ?}'', ``{Which states border Utah ?}'' or ``{Which experimental drugs interact with food ?}'' are tagged with noun labels for the words ``cross'', ``border'' and ``interact'' instead of verb labels.

\begin{table}[ht]
\caption{Test results of the English knowledge base of 90 rules for question analysis on DBpedia and biomedical domains.}
\label{tab:kbEvalResult1}
\begin{tabular}{p{3.75cm}@{\quad} | l | l }
\hline Factor & DBpedia & Biomedical\\
\hline
Successfully processed & 47/50 & 21/25 \\
\hline 
Unknown structure patterns & 0/50 & 0/25 \\
 
Incorrect word segmentation & 3/50 & 3/25 \\

Incorrect part-of-speech tagging & 0/50 & 1/25  \\
\hline
\end{tabular}
\end{table}

To correct the question analysis errors on the two sets of test questions,  we spent 5 further hours to add 31 exception rules to the knowledge base. Finally, in total 12 hours,  we constructed a knowledge base of 90 rules for English question analysis, including the default rule and 89 exception rules. The new evaluation results of question analysis on the  DBpedia and biomedical domains are presented in Table \ref{tab:kbEvalResult1}. 

Table \ref{tab:exceptRuleinEng} shows the number of exception rules in each exception layer of the knowledge base while the number of rules corresponding to each question structure type is presented in Table \ref{tab:numRuleVE}.

\begin{table}[ht]
\caption{Number of exception rules in layers in our English knowledge base.}
\label{tab:exceptRuleinEng}
\begin{tabular}{l@{\quad} l}
\hline Layer & Number of rules \\
\hline 1 & 10 \\
		2 & 21 \\
		3 & 31 \\
		4 & 20 \\
		5 & 7\\
\hline
\end{tabular}
\end{table}

As the intermediate representation in KbQAS is different from AquaLog, it is difficult to directly compare our knowledge acquisition approach with the ad-hoc question analysis approach in AquaLog on the English domain. However, this experiment on English questions shows the abilities to quickly build a new knowledge base and easily adapt to a new domain and a new language.

As illustrated in Section \ref{subsec:kap}, this experiment also presented a process of building a knowledge base for  question analysis without any concept or entity type information. However, we found that the concept or entity type information in noun phrases is useful and  can help to reduce  ambiguities in question structure patterns. When adapting  our knowledge acquisition approach for question analysis to anther target domain (or language), we can simply use the heuristics presented in Section \ref{ssubsec:sa}  and a dictionary to determine whether a noun phrase is a concept or entity type. The dictionary can be (automatically) constructed by extracting concepts from the target domain and their synonyms from available semantic lexicons such as WordNet \cite{FellbaumC98}. 

\subsection{Experiment on answering Vietnamese questions}
\label{subsec:kbqas}

To evaluate the answer retrieval component of KbQAS, we used the ontology modeling the organizational structure of the VNU University of Engineering and  Technology,  as  mentioned in  Section \ref{subsec:illex}, as target domain. This ontology was manually constructed by using the Protégé  platform \cite{Protege2002}.
From the list of 88 questions, as mentioned in Section \ref{sec:qav}, we employed 74 questions which were successfully analyzed by the question analysis component.    % on these questions as mentioned in table \ref{tab:correctResult}.  

\begin{table}[ht]
\caption{Questions successfully answered.}
\label{tab:qsa}
\begin{tabular}{l@{\quad} l}
\hline Type & \# questions \\
\hline

No interaction with users & 30/74\\

With interactions with users & 31/74\\

\hline

Overall & {61/74} (\textbf{82.4\%}) \\

\hline
\end{tabular}

\end{table}

 The performance result is presented in Table \ref{tab:qsa}.  The answer retrieval component produces correct answers for
61 out of 74 questions, obtaining a promising accuracy of 82.4\%. The number of correctly answered questions corresponding to each question structure type can be found in the third column of Table \ref{tab:numRuleVE}. Out of those, 30 questions can be answered automatically without interaction with  users. In addition, 31 questions are correctly answered with the help from the users to handle ambiguity cases, as illustrated in the first example in Section \ref{ssec:arc}.

\begin{table}[ht]
\caption{Questions with unsuccessful answers.}
\label{tab:qusa}
\begin{tabular}{l@{\quad} l}
\hline Type & \# questions\\
\hline
Ontology mapping errors & 6/74 \\
Answer extraction errors &  7/74 \\
\hline
\end{tabular}
\end{table}

Table \ref{tab:qusa} gives the limitations that will be handled in future KbQAS versions.
The  errors raised by the ontology mapping module are due to the
target ontology construction lacking a full domain-specific conceptual coverage and some relationships between concept pairs.  So specific terms or relations in  query tuples cannot be mapped or are
incorrectly mapped to the corresponding elements in the target ontology to produce the  ontology tuples. 
Furthermore, the answer extraction module fails to extract the answers for 7 questions because: (i) Dealing with questions having the   \textit{Compare} question structure involves specific services. For example, handling the question {``sinh viên nào có điểm trung bình cao nhất khoa công nghệ thông tin ?''} (``Which student has the highest grade point average in the faculty of Information Technology ?'') requires a comparison mechanism to rank students according to their GPA. (ii) In terms of four  \textit{Clause} structure questions and one \textit{Affirm\_MoreTuples} structure question for which KbQAS failed to return
answers (see Table \ref{tab:numRuleVE}), combining their sub-questions triggers complex inference tasks and bugs which are difficult to handle in the current KbQAS version. 
 
\section{Conclusion and future work}
\label{sec:conclusion}

In this paper, we described the first ontology-based question answering system for Vietnamese, namely KbQAS. KbQAS contains  two components: natural language question analysis and answer retrieval. The two components are connected by an intermediate representation element capturing the semantic structure of any input question, facilitating the matching process to a target ontology to produce an answer. Experimental results of KbQAS on a wide range of questions are promising. Specifically, the answer retrieval module achieves an accuracy of 82.4\%.

In addition, we proposed a question analysis approach for systematically building a knowledge base of rules 
to convert the input question into an intermediate representation element. Our approach allows for systematic control of interactions between rules and keeping consistency among them. 
We believe that our approach is important especially for
under-resourced languages where
annotated data is not available. Our approach could be
combined nicely with the process of annotating corpora
where, on top of assigning a label or a representation to
a question, the experts just have to add one more rule
to justify their decision. Incrementally,
an annotated corpus and a rule-based system can
be obtained simultaneously.
Furthermore, our approach can be applied to open-domain question answering where the technique requires an analysis to transform an input question into an explicit representation of some sort. Obtaining a question analysis  accuracy of 84.1\%  on Vietnamese questions and taking 12 hours to build a knowledge base of 90 rules for analyzing English questions,  the question analysis experiments show that our approach enables individuals to easily build a new knowledge base or adapt an existing knowledge base to a new domain or a new language. 

In the future, we will extend KbQAS to be an open-domain question answering system which can answer various questions over  Linked Open Data such as DBpedia or YAGO. 
In addition, it would be interesting to investigate the process of building a knowledge base for question analysis,  which directly converts the input questions into queries (e.g. SPARQL queries) on  Linked Open Data.

\section*{Acknowledgments}
  This work is partially supported by the Research Grant No. QG.14.04 from Vietnam National University, Hanoi (VNU). Most of this work was done while the first two authors was at the VNU University of Engineering and Technology.  The first author is supported by an International 
Postgraduate Research Scholarship and a NICTA 
NRPA Top-Up Scholarship.

\section*{Appendix}

\subsection*{A. Definitions of question structure types}
\label{app:dqs}

We define question structures types: \textit{Normal, UnknTerm, UnknRel, Definition, Affirm, ThreeTerm, Affirm\_3Term,  Affirm\_MoreTuples, Compare, And, Or, Combine, Clause} as follows:

$\bullet$ A \textit{Normal} structure question has only one query tuple in which \emph{Term}$_3$ is missing. 

$\bullet$ An \textit{UnknTerm} structure question has only one query tuple in which \emph{Term}$_1$ and \emph{Term}$_3$ are missing.

$\bullet$ An \textit{UnknRel}  structure question has only one query tuple  in which \emph{Relation} and \emph{Term}$_3$ are missing. For example, the question ``{List all the publications in knowledge media institute}'' has one query tuple ({\textit{UnknRel}, \textit{QU-listClass}, publications, ?, knowledge media institute, ?}).

$\bullet$ A \textit{Definition} structure question has only one query tuple which lacks \emph{Term}$_1$, \emph{Relation} and \emph{Term}$_3$. For example, the question ``What are research areas ?'' has one query tuple ({\textit{Definition}, \textit{QU-who-what}, ?, ?, research areas, ?}).

$\bullet$ An \textit{Affirm} structure question  is a question which belongs to one of three  types \textit{Normal}, \textit{UnknRel} and \textit{UnknTerm}, and  has  the \textit{YesNo} question category.
 For example, the question ``Is Tran Binh Giang a PhD student ?'' has the \textit{Affirm} question structure and one query tuple ({\textit{UnknRel}, \textit{YesNo}, PhD student, ?, Tran Binh Giang, ?}).

$\bullet$ A \textit{ThreeTerm} structure question has only one query tuple where \emph{Term}$_1$ or \emph{Relation} could be missing. An example for this structure type is illustrated in Figure \ref{fig:kbqa}.

$\bullet$ An \textit{Affirm\_3Term} structure question is the question which belongs to  \textit{ThreeTerm} and has the  \textit{YesNo} question category. 
For example, the question {``số lượng sinh viên học lớp K50 khoa học máy tính là 45 phải không ?''} ({``45 is the number of students enrolled in the K50 computer science
course, is it not ?''}) has the \textit{Affirm\_3Term} question structure and one query tuple ({\textit{ThreeTerm}, \textit{ManyClass}, sinh viên\textsubscript{student}, học\textsubscript{enrolled}, lớp K50 khoa học máy tính\textsubscript{K50 computer science course}, 45}).% 

$\bullet$ An \textit{Affirm\_MoreTuples} structure question has more than one query tuple and belongs to the \textit{YesNo} question category. For example, the question {``tồn tại sinh viên có quê ở Hà Tây và học khoa toán phải không ?''} (``Is there some student enrolled in the faculty of Mathematics, whose hometown is Hatay ?'') has the \textit{Affirm\_MoreTuples} question structure and  two query tuples ({\textit{Normal}, \textit{YesNo}, sinh viên\textsubscript{student}, có quê\textsubscript{have\ hometown}, Hà Tây\textsubscript{Hatay}, ?}) and ({\textit{Normal}, \textit{YesNo}, sinh viên\textsubscript{student}, học\textsubscript{enrolled}, khoa Toán\textsubscript{faculty\ of\ Mathematics}, ?}).

$\bullet$ A \textit{Compare} structure question is a question which belongs to one of three  types \textit{Normal}, \textit{UnknRel} and \textit{UnknTerm}, and it contains a comparison phrase which is detected by the preprocessing module. In this case, \emph{Term}$_3$  is used to hold the comparison information. 
For example, the question {``sinh viên nào có điểm trung bình cao nhất khoa công nghệ thông tin ?''} ({``Which student has the highest grade point average in the
faculty of Information Technology ?''}) 
has  the \textit{Compare} question structure and one query tuple ({\textit{Normal}, \textit{Entity}, sinh viên\textsubscript{student}, điểm trung bình\textsubscript{grade\ point\ average}, khoa công nghệ thông tin\textsubscript{faculty\ of\ Information\ Technology}, cao nhất\textsubscript{highest}}).

$\bullet$ An \textit{And} or \textit{Or} structure question contains the word {``mà\textsubscript{and}''} ({``và\textsubscript{and}''}) or {``hoặc\textsubscript{or}''}, respectively, and it has more than one query tuple (i.e. two or more sub-questions). The \textit{And} type returns the final answer as an intersection (i.e. overlap) of the answers  of the sub-questions, while the \textit{Or} type returns the final answer as an union of the answers  for the sub-questions.

For example, the question ``Which projects are about ontologies and the semantic web ?'' has the \textit{And} question structure  and two query tuples ({\textit{UnknRel}, \textit{QU-whichClass}, projects, ?, ontologies, ?}) and ({\textit{UnknRel}, \textit{QU-whichClass}, projects, ?, semantic web, ?}). 

The question ``Which publications are in knowledge media institute related to compendium ?'' has the \textit{And} question structure  and two 
query tuples ({\textit{UnknRel}, \textit{QU-whichClass}, publications, ?, knowledge media institute, ?}) and ({\textit{Normal}, \textit{QU-whichClass}, publications, related to, compendium, ?}). 

The question ``Who is interested in ontologies or in the semantic web ?'' has the \textit{Or} question structure and two query tuples ({\textit{UnknTerm}, \textit{QU-who-what}, ?, interested, ontologies, ?}) and ({\textit{UnknTerm}, \textit{QU-who-what}, ?, interested, semantic web, ?}).

However,  such question as  {``Phạm Đức Đăng học trường đại học nào và  được hướng dẫn bởi ai ?''} ({``Which university does Pham Duc Dang enroll in and who tutors him ?''})  contains {``và\textsubscript{and}''}, but it will have the  \textit{Or} question structure and two query tuples   ({\textit{Normal}, \textit{Entity}, trường đại học\textsubscript{university}, học\textsubscript{enroll}, Phạm Đức Đăng\textsubscript{Pham\ Duc\ Dang}, ?}) 
 and  
({\textit{UnknTerm}, \textit{Who}, ?, hướng dẫn\textsubscript{tutor}, Phạm Đức Đăng\textsubscript{Pham\ Duc\ Dang}, ?}).

$\bullet$ A \textit{Combine} structure question is constructed from two or more independent sub-questions. Unlike the Or structure type, the  query tuples  in the \textit{Combine} type do not share the same term or \emph{Relation}. For example, the question {``Ai có quê quán ở Hà  Tây và ai học khoa công nghệ thông tin ?''} (``Who has hometown of Hatay, and who enrolls in the faculty of Information Technology ?'') has the  \textit{Combine}  question structure  and two query tuples  
({\textit{UnknTerm}, \textit{Who}, ?, có quê quán\textsubscript{has\ hometown}, Hà  Tây\textsubscript{Hatay}, ?}) 
and
({\textit{UnknTerm}, \textit{Who}, ?, học\textsubscript{enroll}, khoa công nghệ thông tin\textsubscript{faculty\ of Information\ Technology}, ?}). 

$\bullet$ A \textit{Clause} structure question has two query tuples, where the answer returned for the second query tuple indicates the missing \emph{Term}$_2$ attribute in the first query tuple. 
For example, the question {``số lượng sinh viên học lớp K50 khoa học máy tính lớn hơn 45 phải không ?''}\footnote{This is the case of our system failing to correctly analyze due to an unknown structure pattern.} (``The number of students enrolled in  K50 computer science course is higher than 45, is it not ?'') has the  \textit{Clause}  question structure and two query tuples ({\textit{Compare}, \textit{YesNo}, 45, ?, ?, lớn hơn\textsubscript{higher\ than}}) and ({\textit{Normal}, \textit{ManyClass}, sinh viên\textsubscript{student}, học\textsubscript{enrolled}, lớp K50 khoa học máy tính\textsubscript{K50\ computer\ science\ course}, ?}). Another example of this \textit{Clause} structure is presented in Section \ref{sssection:qam}.

%; \textit{Term$_2$} and \textit{Term$_3$}, if exist, represent entities$/$instances (maybe representing concepts in case of \textit{Definition} question structure).
%\subsection*{B. Definitions of question category}

\vspace{7pt}
\noindent In general, \emph{Term}$_1$ represents a concept, excluding cases of  \textit{Affirm}, \textit{Affirm\_3Term} and \textit{Affirm\_MoreTuples}. In addition, {\emph{Term}$_2$} and {\emph{Term}$_3$} represent entities (i.e. objects or instances), excluding the cases of  \textit{Definition} and \textit{Compare}.

\subsection*{B. Definitions of Vietnamese question categories}
\label{app:dqt}

In KbQAS,  a question
is classified as one of the following classes: \textit{HowWhy, YesNo, What, When, Where, Who, Many, ManyClass, List}, and \textit{Entity}.
To identify question categories, we specify a number of JAPE grammars using the \textit{NounPhrase} annotations and the {question-word} information given by the preprocessing module.% Obviously using this method in question-phrases detection phase will result in ambiguity when a question belongs to multiple categories.% We allow for this and resolve the ambiguity in the semantic analysis module.

$\bullet$ A \textit{HowWhy}-category question refers to a cause  or a method, containing a \textit{TokenVn } annotation covering such strings as {``tại sao\textsubscript{why}''}  or {``vì sao\textsubscript{why}''}  or  {``thế nào\textsubscript{how}''}  or {``là như thế nào\textsubscript{how}''}. This is similar to \textit{Why}-questions or \textit{How is/are} questions in English. 

$\bullet$ A \textit{YesNo}-category question requires a true or false answer, containing  a \textit{TokenVn } annotation covering such strings  as  {``có đúng là\textsubscript{is\ that}''}  or {``đúng không\textsubscript{are \ those}''}  or {``phải không\textsubscript{are \ there}''} or {``có phải là\textsubscript{is \ this}''}.

$\bullet$ A \textit{What}-category question contains  a \textit{TokenVn } annotation covering such strings  as {``cái gì\textsubscript{what}''}  or {``là gì\textsubscript{what}''}  or {``là những cái gì\textsubscript{what}''}.  This question type is similar to \textit{What is/are} questions  in English. 

$\bullet$ A \textit{When}-category question contains  a \textit{TokenVn } annotation covering such strings  as {``khi nào\textsubscript{when}''}  or {``vào thời gian nào\textsubscript{which \ time}''}  or {``lúc nào\textsubscript{when}''}  or {``ngày nào\textsubscript{which\ date}''}.

$\bullet$  A \textit{Where}-category question contains  a \textit{TokenVn } annotation covering such strings  as {``ở nơi nào\textsubscript{where}''} or {``là ở nơi đâu\textsubscript{where}''} or {``ở chỗ nào\textsubscript{where}''}.

$\bullet$  A \textit{Who}-category question contains  a \textit{TokenVn } annotation covering such strings  as {``là những ai\textsubscript{who}''}  or {``là người nào\textsubscript{who}''}  or {``những ai\textsubscript{who}''}.

$\bullet$  A \textit{Many}-category question contains  a \textit{TokenVn } annotation covering such strings  as   {``số lượng\textsubscript{how\ many}''}  or {``là bao nhiêu\textsubscript{how\ much/many}''} or {``bao nhiêu\textsubscript{how much/} \textsubscript{many}''}. This question type is similar to \textit{How much/many is/are} questions  in English.

$\bullet$ A \textit{ManyClass}-category question contains  a \textit{TokenVn } annotation covering such strings  as   {``số lượng\textsubscript{how} \textsubscript{many}''}  or {``là bao nhiêu\textsubscript{how} \textsubscript{much/many}''} or {``bao nhiêu\textsubscript{how} \textsubscript{much/many}''},  followed by a \textit{NounPhrase} annotation. This type is similar to   \textit{How many NounPhrase}-questions in English.

$\bullet$  An \textit{Entity}-category question contains a \textit{NounPhrase} annotation  followed by a \textit{TokenVn } annotation covering such strings  as   {``nào\textsubscript{which}''} or {``gì\textsubscript{what}''}. This type is similar to   \textit{which/what NounPhrase}-questions  in English. 

$\bullet$ A \textit{List}-category question contains  a \textit{TokenVn } annotation covering such strings  as   {``cho biết\textsubscript{give}''} or {``chỉ ra\textsubscript{show}''} or {``kể ra\textsubscript{tell}''}, or {``tìm\textsubscript{find}''} or {``liệt kê\textsubscript{list}''},  followed by a \textit{NounPhrase} annotation.

\bibliographystyle{abbrvnat}
%\balance
\bibliography{references}

\end{document}